*The Impact of Post-editing and Machine Translation on Creativity and Reading Experience*

Ana Guerberof Arenas

Antonio Toral

University of Surrey/University of Groningen

University of Groningen

*Abstract*

This article presents the results of a study involving the translation of a fictional story from English into Catalan in three modalities: machine-translated (MT), post-edited (MTPE) and translated without aid (HT). Each translation was analysed to evaluate its creativity. Subsequently, a cohort of 88 Catalan participants read the story in a randomly assigned modality and completed a survey. The results show that HT presented a higher creativity score if compared to MTPE and MT. HT also ranked higher in narrative engagement, and translation reception, while MTPE ranked marginally higher in enjoyment. HT and MTPE show no statistically significant differences in any category, whereas MT does in all variables tested. We conclude that creativity is highest when professional translators intervene in the process, especially when working without any aid. We hypothesize that creativity in translation could be the factor that enhances reading engagement and the reception of translated literary texts.

*Introduction*

Artificial Intelligence and Machine Translation (MT) are at the forefront of technological advances and are becoming ubiquitous in society (Bojar et al. 2016; Vaswani et al. 2017). In the media we read time and again that MT will soon substitute for professional translators. In this desolate prospective landscape, creativity is constantly referred to as the characteristic that differentiates humans from machines. The reality of MT for more creative texts might not be as propitious as first appears, however. Research in Natural Language Processing (NLP) has tested the usability of MT for literary texts (Toral, Wieling, and Way 2018), showing that MT post-editing might help literary translators, for example when it comes to productivity. At the same time, however, the perception of translators is that the more creative the literary text, the less useful MT post-editing is (Moorkens et al. 2018). In the opinion of many professional translators, MT output needs to be adequate for its purpose and this might not be the case when creativity is considered (Cadwell et al. 2016, 245).

To our knowledge however, there has been no attempt to quantify creativity in different translation modalities involving MT. Further, and since the main aim of the translation of a literary text is presumably to maintain the reading experience of the original, we deem it relevant to investigate the reader's experience when faced with literary texts translated by or with the aid of MT, a topic that has not been studied to date. Against this background, we present the results of a study (part of a larger project)[1] that tests a methodology designed for exploring creativity in different translation modalities, and to capture aspects of the user experience when reading in these modalities. Below we first present work related to our study, before outlining our novel methodology and presenting our results.

*Related Work*

Creativity is an area that has received limited attention in Translation Studies (TS). One reason for this could be that creativity is a complex concept to define and quantify (Rojo 2017, 351–52; Malmkjær 2019, 14), but it is also the case that translators are not usually regarded as creators (Hewson 2016, 11) and translation is perceived as uncreative, thus its derivative copyright status (Venuti 1998, 50). Perhaps, this is also because creativity has traditionally been associated with a talent that one is born with and that cannot be learnt, understood or quantified; therefore, as an innate characteristic, only a chosen few can instinctively put it into practice without being able to explain it, let alone understand it themselves. However, this pre-conceived notion is receding and creativity is increasingly regarded as an ability that is inherent to all humans, and not limited to those working in the arts, and one that can be trained and applied in a variety of domains including translation (Malmkjær 2019, 31).

---

[1] https://cordis.europa.eu/project/id/890697

Kussmaul (1995; 1991; 2000a; 2000b), in his seminal work on creativity, analyses creative techniques found in literary translations carried out by students and compares them to those used by professional translators with the goal of understanding the process of translating creatively. He uses Think Aloud Protocols to collect data (Kussmaul and Tirkkonen-Condit 2007). In his attempt to operationalize creativity, he introduces the concept of scenes and frames in the source text that constitute areas that require the translator to think creatively to find solutions that are out of the ordinary, going beyond the routine. Kussmaul posits that for a translation to be creative it has to be novel and acceptable; in other words, it offers a new solution in the target text and creates a meaning that it is judged to be correct by experts.

The most comprehensive attempt to operationalize and quantify creativity in translation to date has been made within the TransComp research framework (Göpferich et al. 2011; Göpferich 2013), and more specifically by Bayer-Hohenwarter (2009; 2010; 2011; 2013). Based on the nine creativity dimensions defined in psychology by Guildford (1950), paired with the work done in TS by Kussmaul, she defines creativity as involving a combination of four dimensions: Acceptability (the translation meets the requirements of the brief), Flexibility (the use of *creative shifts,* a target text that departs from the linguistic structure of the source text as opposed to a *reproduction* that represents a literal rendering of the source text), Novelty (how unique a translation is in comparison to others) and Fluency (the number of translation solutions provided for one problem by one translator). She then develops a scoring system to rate translations with a view to understanding competence and training in translation.

Hewson (2016) analyses the stages in creative translation based on the existing literature on creativity and TS and with the aim of offering a way to teach creativity in the translation classroom, not only for those texts that are a priori classified as "creative" but for all texts. He explains the difficulties that trainees have when confronting a new text and offers a broad definition of creativity, which involves the process of paraphrasing and rewriting in both the ST and TT within the boundaries of interpretation of the ST.

Rojo and colleagues (Rojo and Ramos Caro 2016; Rojo and Meseguer 2018; Rojo and Ramos Caro 2018) explore creativity tangentially, i.e. they look at personality traits, emotion, and expertise in relation to creativity in translation on the basis that a creative translation is a sign of a proficient translation. Although they find no direct correlation between the number of mistakes made by translators and their creativity traits, they do find a correlation between creative personalities and creative translations (analysed using the *creative shifts* described by Bayer-Hohenwarter (2011) and discussed above). However, this correlation is weak to moderate. They also find that positive feedback and constructive criticism in the classroom might strengthen students' creativity, but that negative feedback also plays a role in generating proficient translations.

Recently, Vieira and colleagues (2020) have explored how commercial CAT tools impact translators working on literary texts in two modalities (HT and MTPE). In the language combination English to Chinese, they find that translators have fewer typing pauses and less keyboard use in MTPE, but there are no differences from HT in time. They also find that translators working with paragraph-level segmentation work faster, have fewer typing pauses, and make less use of the keyboard. Finally, and most relevant to our work, they find that creativity, according to five expert judges rating all the target translations, is no different depending on the translation modality, although there is little or no agreement between experts. Interestingly, the translators reported having changed the text substantially when post-editing due to the unsatisfactory quality of the MT output.

Some researchers have also explored the use of technology to understand translation styles in literary texts (Kenny 2000; Saldanha 2005), to compare translations, or even to explore the translator's own draft versions (Youdale 2019), using tools that allow them to see patterns not only in the source text (for example, patterns related to sentence length or word frequency), but also in the target text, and to take a "distant" view of the text and the translation, one that differs from the translator's normal view, which is more immersed in the actual translation. More recently, the translator's voice has been explored, with Kenny and Winters (2020) analysing the translations of Hans-Christian Oeser in two versions (the original translation, created without aid, and the post-edited version), and finding that post-editing has the effect of diminishing his voice in literary translation.

Although it is widely acknowledged that MT is a more suitable tool for technical texts because of their repetitive nature and the need for accuracy, some researchers in NLP have also shown an interest in testing the applicability of MT to literary translation. For example, Besacier and Schwartz (2015) explore the feasibility of using a Statistical MT (SMT) system to translate a literary text from English to French. In this experiment, the MT output is then post-edited by non-professional translators and read by a group of readers. Although the researchers report a perceived decrease in quality, they argue that MT can reduce translation time and provide faster access to different language markets for authors, as well as a possible aid for second language reading.

Toral and colleagues (Toral and Way 2015a; 2015b; Toral, Wieling, and Way 2018) explore the possibility of building customized MT engines to translate literary work from Spanish to Catalan, and also compare the use of Neural MT (NMT) and SMT in a literary context. Their results show that NMT outperforms SMT consistently and by a wide margin. NMT is also the preferred paradigm for the translators who participate in related experiments in which MT is used as a translation aid, although these same translators clearly prefer to translate without MT, as they feel more creative when working on their own (Moorkens et al. 2018).

Matusov (2019) reports on an NMT system tailored for translating literature from English to Russian and from German to English. This system leads to better automatic evaluation metric scores than those achieved with general domain NMT. Similarly, Kuzman and colleagues (2019) report on a system tailored to translate novels from English to Slovenian. While the quality evaluation results are worse than those from Google Translate, the authors report two interesting findings: a) gains in productivity when using MT as opposed to translating without it and b) the system tailored to a given author obtained promising results.

In studies that explore user experience of MT and MTPE, research methods have involved monitoring users using an eye-tracker while they perform different tasks and then analysing their task completion, the time taken to complete tasks and their satisfaction when exposed to different translation modalities (Doherty and O'Brien 2014; Castilho 2016; Guerberof, Moorkens, and O'Brien 2019). Participants have also been asked to choose preferred translations (Daems and Macken 2019). The findings encourage the use of MT and post-editing in technical settings. However, these methods might be insufficient to measure user experience when MT is applied to literary texts because that experience is expected to occur at a more abstract or emotional level.

Since the interest in the present study is to see how users are "engaged" or "enjoy" a literary text translated using different modalities, we turn to psychology and literary studies. Transportation theory (Green and Brock 2000) looks at how people are transported into a narrative world in works of fiction or non-fiction to determine how reading might impact people's behaviour, beliefs or attitudes, i.e. how being lost in a story (transported) can change beliefs about issues entailed in the story. Transportation involves imagery, emotional response and attentional focus, and it has been found to be strongly correlated with enjoyment (Green, Brock, and Kaufman 2004). Interestingly, stylistic techniques or literary devices help readers to see aspects of human experience differently, therefore well-crafted literary texts lead to higher transportation and enjoyment levels. When readers are disrupted by having to do another task (Green and Brock 2000) or if the text contains errors, attentional focus decreases, hence enjoyment and transportation decrease (Dixon et al. 1993; Hakemulder 2004).

To analyse this phenomenon, scales have been developed in recent years to measure transportation in media, film and literature. These scales are known generically as narrative engagement scales (Busselle and Bilandzic 2009; Kuijpers 2014). These scales, used and tested in psychology and applied also to literary studies, can help to determine if readers have different engagement levels when different translation modalities are applied; that is, to test if there is an 'estrangement' (e.g. loss of attentional focus) effect that might prevent readers from enjoying a text translated using MT as much as a text translated without any aid.

*Research Methodology*

The overall question driving our research is *How does creativity in different translation modalities (MT, MTPE and HT) impact the reading experience?* This question possibly merits several experimental projects; for this initial study the focus is on these two sub-questions:

**RQ1:** Can we quantify the creativity in texts translated by humans as opposed to those produced with the aid of machines?

**RQ2:** Do users reading translated material produced using different translation modalities have different reading experiences?

In the following sections, the different aspects of the methodology applied to explore the answers to these questions are described.

*Source Text*

The story *Murder in the Mall* by Sherwin B. Nuland (1994) containing 2,277 words was translated using the three modalities proposed (HT, MT and MTPE). This text has already been used in previous research applying narrative engagement scales (Green and Brock 2000), so it was considered appropriate to measure engagement. It was also considered suitable for an initial experiment using MT as in this story, action and emotion prevail over style. We felt that for an initial test, it was futile to use a particularly challenging text for MT. We used an adapted version (Mangen and Kuiken 2014), which describes a tragic event in Connecticut: A psychiatric patient, trying to kill a nine-year old girl, stabs a homeless person to death who has instinctively intervened to protect the child; the event is witnessed and retold in part by the girl's mother.

*Translation process*

The text was translated from English into Catalan. We chose this translation direction for two reasons: our interest and knowledge of Catalan, and the availability of an MT engine customized for literary texts available for this language combination. The HT and MTPE versions were provided by two professional translators who specialize in literary translation.

The translators used the PET post-editing tool (Aziz, Castilho, and Specia 2012), and the text was segmented by paragraph, based on our translation experience and following previous research (Toral, Wieling, and Way 2018; Vieira et al. 2020). To reduce the effect of the translator in the experiment, i.e. the risk that a reader would engage more with one translator's work because they preferred that particular translator's style (Saldanha 2005), each professional translated and post-edited 50% of the text, and then the text was aggregated so, in fact, each of the two translation modalities contained the aggregated translations of the two translators. Although literary translators tend to translate on their own, it is not altogether new to collaborate with others to create a translation (Cordingley and Manning 2016). Finally, the MT modality was based on the output of a state-of-the-art system based on the Transformer neural network architecture (Vaswani et al. 2017) trained with more than 130 novels in English and their translations into the target language, as well as more than 1000 books in the target language (Toral, Oliver, and Ribas-Bellestín 2020).

*Translators*

The two professional translators, one female (Translator 1), and one male (Translator 2) had 4- and 27 years' experience translating literary texts and had translated 15 and 18 novels respectively during this time. After finishing the assignment, a short Google Forms questionnaire was used to gather the translators' opinions on the quality of the MT output. Table 1 shows the results of the questionnaire.

| Question | Translator 1 | Translator 2 |
|---|---|---|
| **How would you rate the MT output for this task?** **1 very unhelpful; 7 Very helpful** | **4** | **4** |
| What specific language related issues cause the most serious and frequent problems? Please, select all that apply. | Word order Polisemy Figure and abstract language Format Mistranslation | Word order Polisemy Gender concordance Mistranslation |
| How much of the meaning expressed in the source text is represented in the translation provided by MT? 1 None; 7 All | 5 | 6 |
| To what extent are the translations fluent, i.e. well-formed grammatically, correct spelling, adhering of common use of terms, intuitively acceptable and could be sensibly interpreted by a native speaker? 1 Incomprehensible; 7 Flawless | 4 | 5 |
| Please, rank the translation method according to speed. (1 the slowest, 7 the fastest) [Translating from scratch] | 6 | 4 |
| Please, rank the translation method according to speed. (1 the slowest, 7 the fastest) [Post-editing MT] | 4 | 5 |
| Please, rank the translation method according to the effort required from you (1 the most effort, 7 the least effort) [Translating from scratch] | 7 | 3 |
| Please, rank the translation method according to the effort required from you (1 the most effort, 7 the least effort) [Post-editing MT] | 4 | 4 |
| Please, rank the translation method according to your preference (7 the most preferred, 1 the least preferred) [Translating from scratch] | 7 | 7 |
| Please, rank the translation method according to your preference (7 the most preferred, 1 the least preferred) [Post-editing MT] | 5 | 4 |

Table 1: Translators' opinions on the MT output

When asked their opinions about post-editing in comparison to translation, they said.
Translator 1:
> In the post-editing phase, I felt that my creativity was limited, and I found it more difficult to think outside the box when I already had a translation provided. I felt a bit uncomfortable 'fixing' the text instead of giving my own translation.

Translator 2
> Rewriting is always perceived as being less creative than translating from scratch. The translator feels the text as not fully their own. Moreover, after post-editing one is often under the impression that the outcome is not the real thing, is not the best that could be offered to a potential reader. It is just not bad, relatively satisfactory, but not the best one could do.

Further to this, the final translation was analysed using HTER, Human-Targeted Translation Error Rate (Snover et al. 2006), an automatic score that reflects the number of edits performed on the MT output normalized by the number of words in the sentence. Research has shown that this metric correlates well with actual post-editing effort (O'Brien, 2011; Guerberof-Arenas, 2012). The closer HTER is to 0 (the lowest possible value), the lower the translator's effort because fewer changes are required. Table 2 shows the HTER results using the MTPE and HT modalities as reference and MT as the hypothesis.[2] Table 3 shows the type of post-edits made by the two translators by percentage.

| Translator | MTPE | HT |
|---|---|---|

---

[2] The reference (HT) is the translated text (sentence) that the hypothesis (MT) is compared against.

| | | |
|---|---|---|
| T1 | 0.34 | 0.54 |
| T2 | 0.31 | 0.59 |

Table 2: HTER results for MTPE and HT

| Translator | Insertions | Deletions | Substitutions | Shifts |
|---|---|---|---|---|
| T1 | 13% | 14% | 60% | 12% |
| T2 | 16% | 17% | 56% | 10% |

Table 3: Percentage of edits in MTPE according to HTER

From the results, we can see that the two translators show similar behaviour when post-editing if we consider the number and type of changes, but also that the MTPE modality was, as expected, closer to the MT than its HT counterpart. The HTER scores for MTPE are also an indication of the quality of the output, as these values are considered to be acceptable for post-editing in the language industry (Schmidtke and Groves 2019).

*Measuring Creativity*

As we saw earlier, in TS there is agreement that a creative product must involve two key aspects: novelty and usefulness (or novelty and adequacy as in Rojo 2017, 353). Novelty refers to new solutions in the target text for problems posed by the original text; and usefulness to a translation that suits its purpose, that it is acceptable in the intended communication context. As Hewson explains, creativity involves producing "unpredictable micro-level translation solutions that are coherent with the macro-level interpretation…" (Hewson 2016, 20), that is, translators should master both languages and offer new solutions that are appropriate to the source text context. In this study, we apply Bayer-Hohenwarter's model (2010) of creativity, albeit modified since it was not possible to compare several translations for one source text (ST) as in a classroom. This model constitutes the first attempt to measure creativity and has become standard in similar studies (as explained in the Related Work section) and it offers a systematic way of measuring creativity in translation which connected well with our research goal. Therefore, we use Acceptability, operationalized by counting the number of errors in the translated texts, and Flexibility, operationalized by analysing and quantifying primary creative shifts in the target text (TT). In our case, the Flexibility and Novelty dimensions are merged into one: we consider the number of creative solutions (shifts) provided for a given problem in the ST, it will be referred to as Novelty from this point onwards.

*Acceptability:* To rate the translated texts we used the harmonized DQF-MQM Framework,[3] which unifies evaluation practices from academia and industry. Because of our definition of acceptability (which refers to the quality of being error free), it was decided to use the Harmonized Error Typology,[4] which classifies errors according to the following high-level error types: Accuracy, Fluency, Terminology, Style, Design, Locale Convention, Verity and Other. For this research, we were interested in the number of errors rather than in the severity of these errors because we are aware that classifying severity is difficult for some reviewers due to subjective interpretation of annotation guidelines, and we wanted to simplify the classification as much as possible for the reviewer.

*Novelty.* To explore novelty, 48 sentences were marked in the English ST as units of high creativity potential (units that either contain creative elements or that required creative solutions in the translation). According to Bayer-Hohenwarter, high-creativity units are simply defined as "problematic units that are deemed to require high problem-solving capacity" as opposed to those that are regarded as "fairly unproblematic, at least for the experienced translators" and that are considered routine units that are standard in the translation practice in question (Bayer-Hohenwarter 2011, 642). This definition seems vague, but for most TS scholars creativity starts with a problem that is not easy to solve for the experienced translator, but that also stems "from the translator's desire to go beyond the fairly standard form adopted during the preliminary draft of the target text" (Hewson 2016, 13). Examples from our data include the following: "the pupils dilate and forever lose their watchful light", "Even through the haze of her frozen incomprehension", "To the locals, Bugsy was known to be a harmless, quiet, good-hearted and even well-behaved man", "driving his cart into him". In general, the units contain comparisons, metaphorical expressions, imagery and abstraction, idioms, verbal phrases or complex syntactic structures.

One of the researchers, also an experienced translator, marked these 48 units. Further to this, an award-winning literary translator from English and German into Catalan with 36 years' experience and with more than 50 literary translations to their name evaluated acceptability and classified each of these creativity units in the target texts (TT) according to the following criteria.

---

[3] https://www.taus.net/qt21-project
[4] https://www.taus.net/qt21-project#harmonized-error-typology

**Reproduction:** All translations that reproduce the ST with the same idea or image, even if they are acceptable.

**Creative shifts (CS):** All translations that deviate from the ST are considered CS as follows:

- <u>Abstraction</u> refers to instances when translators use more vague, general or abstract TT solutions as compared with the ST. A more generic element is used.
- <u>Concretisation</u> refers to instances when the TT evokes a more explicit, more detailed and more precise idea or image than the ST. A more concrete element is used.
- <u>Modification</u> refers to instances when translators use the same level of abstraction (e.g. express a ST metaphor with a different TT metaphor without the image becoming more abstract or concrete).

We are aware that more than one expert reviewer would be desirable because of the subjective nature of these classifications and the usual divergence between reviewers (Guerberof-Arenas 2017; Vieira et al. 2020); however, the excellent credentials of the reviewer and the exploratory nature of this study justified this choice.

*Creativity Scoring.* To calculate a final creativity score , the following formula, based on Bayer-Hohenwarter (2010), is used: a creative shift is worth one point divided the number of units of creativity potential to obtain a percentage of CSs; the errors, found in the target text divided by the number of total words in that text, are penalized by subtracting them from the CSs percentage. Therefore, the two concepts that we use to define creativity (acceptability and novelty) are transformed into a single number. The result is multiplied by 100 to express it as a percentage.

$$creativity\ score\ = (\#CS \div units) - (\#\ errors \div \#\ words\ in\ TT) * 100$$

*Measuring User Experience: Narrative Engagement, Enjoyment and Translation Reception*

An on-line questionnaire consisting of different sections (Figure 1) was distributed to participants using Qualtrics software.[5]

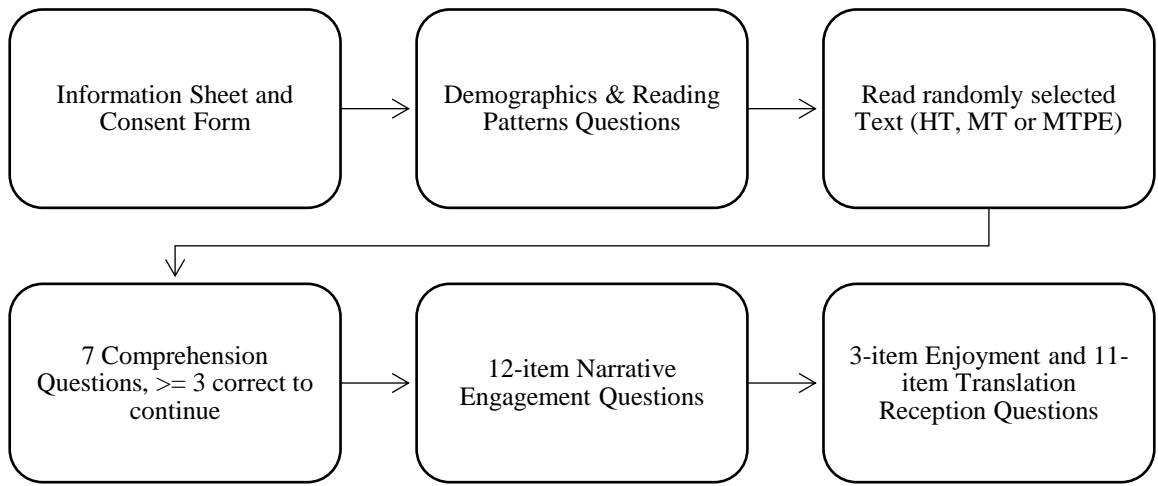

Figure 1 Questionnaire Workflow

*Demographics and Reading Patterns:* This section of the questionnaire contains questions on demographics and reading patterns (e.g. "What genre do you usually read? How often have you read in Catalan in the last 12 months?").

*Comprehension Questions:* After reading the text, the participants answered seven 4-choice questions to ensure basic comprehension of the story. The participant could continue only if they answered at least three questions correctly.

*Narrative Engagement:* Participants were presented with a 12-item Narrative Engagement questionnaire (Busselle and Bilandzic 2009) that the participants were asked to respond in a 7-point Likert scale. The questionnaire focuses on four categories: Narrative understanding (e.g. "At points, I had a hard time making sense of what was going on in the story"), Attentional focus (e.g. "While reading, I found myself thinking about other things"), Narrative presence (e.g. "The story created a new world, and then that world suddenly disappeared when the story ended"), and Emotional engagement (e.g. "I felt sorry for some of the characters in the story").

---

[5] The questionnaire and the data are available at https://github.com/antot/pemt_creativity_reading.

*Enjoyment and Translation Reception.* Participants were then asked to answer questions designed to address enjoyment: "How much did you enjoy reading the text?", "Do you think this text is an example of good literature?", "Would you recommend this text to a friend?". These questions were borrowed from experimental research on the effects of foregrounding[6] in relation to specific text qualities (Dixon et al. 1993; Hakemulder 2004).

To our knowledge, there is no existing scale to measure Translation Reception. Therefore, we devised four questions to measure this variable: "How easy was the text to understand?", "What do you think of the translation?", "How would you like to read a text by the same author and translator?" and "How would you like to read a text by the same author but a different translator?". The participants were also asked eleven other questions related to translation (e.g. "Did you notice you were reading a translation?"). The participants were asked to use a 7-point Likert-type scale or closed questions (Yes, No, I don't know).

*Participants.* The criteria for the inclusion of participants was that they were native Catalan speakers and 18 years or older. To incentivize participation, readers were offered 10-euro compensation for their time. To recruit them, we decided to target forums for readers and not to use our own social media channels to avoid any bias. We posted ads in the following places: Goodreads (Lectura en català Group[7]), Relats en català (a web forum for readers and writers in Catalan),[8] Catalans in Ireland,[9] Catalans in Holland,[10] Catalans in the UK[11] (Facebook groups) and the Facebook[12] and Twitter accounts [13] of the Llengua Catalana Departament of the Catalan Government (with over thirty thousand followers). We posted the ad at different intervals during the time the questionnaire was active, namely from April 8th to May 30th 2020.

---

[6] In literary studies, foregrounding refers to the effect of certain language features that serve to change the attention of the reader.
[7] https://www.goodreads.com/group/show/61003-lectura-en-catal
[8] http://relatsencatala.cat/
[9] https://www.facebook.com/groups/CatalansIrlanda
[10] https://www.facebook.com/groups/203190096419372
[11] https://www.facebook.com/groups/catalansuk
[12] https://www.facebook.com/llenguacatalana/
[13] @llenguacatalana

101 participants completed the questionnaire. From this cohort, we eliminated those who had filled it out more than once (participants with the same IP address and the same email account). The final number of participants was 88. Table 4 contains demographic information on participants.

| Variable | Categories in percentage | | |
|---|---|---|---|
| **Gender** | Women | Men | Prefer not to say |
| | 50 | 37 | 1 |
| **Age** | 18-34 | 35-54 | 55-84 |
| | 58 | 21 | 9 |
| **Mother Tongue** | Catalan | Catalan and Spanish | |
| | 50 | 38 | |
| **Profession** | Related to language | Other | N/S |
| | 21 | 62 | 5 |

Table 4 Participants Demographics

There were four questions that addressed reading patterns (frequency, reading enjoyment, reading in the last 12 months, and time spent on reading). There were no statistically significant differences in reading patterns in the three modalities (HT M= 4.08; MTPE M = 4.11; MT M = 4.22).

*Results*

This section presents the results for Creativity and User experience in turn.

*Results on Creativity*

*Acceptability.* Table 5 summarizes the results from the acceptability evaluation (only the criteria where errors were found are shown).

| Analysis Criteria | # errors in HT | # errors in MTPE | # errors in MT |
|---|---|---|---|
| Accuracy | 18 | 9 | 45 |
| Fluency | 8 | 14 | 55 |
| Terminology | 1 | 0 | 0 |
| Style | 22 | 22 | 73 |
| Total | **49** | **45** | **173** |

Table 5: Acceptability according to translation modality and analysis criteria

The most obvious (but also expected) result is that the MT contains more errors than the other two modalities put together. In fact, the MT contains 3.53 times more errors than HT and 3.84 times more than MTPE.

Looking at types of errors, the results on fluency and on accuracy of HT and MTPE deserve further discussion. MTPE scores best for accuracy, i.e., presents fewest accuracy errors. However, when it comes to fluency, HT has the lowest number of errors; it seems that accuracy might be 'sacrificed' slightly in HT to obtain a more fluent translation, possibly because the translators are dealing with more expressive elements of the TT. Coincidentally, when evaluating outputs in our experiment, the translators also scored the accuracy of this MT output higher than the fluency. These results seem in line with the view often expressed by translators that they are constrained by MT proposals, which makes MTPE texts less fluent (Moorkens et al. 2018). They are also consistent with previous research that indicates that the final quality of post-edited products (in terms of number of errors) is equal to or higher than that of translations produced without any aid (Guerberof-Arenas 2012).

*Creative Shifts.* Table 6 shows the results for reproduction and CS in the 48 creative units as classified by the independent reviewer.

| TM | Reproduction | | CS | | | | | | Total CS | | Invalid | |
|---|---|---|---|---|---|---|---|---|---|---|---|---|
| | | | Abstraction | | Modification | | Concretisation | | | | | |
| | Count | % | Count | % | Count | % | Count | % | Count | % | Count | % |
| HT | 13 | **27.1** | 12 | 25 | 5 | 10.4 | 17 | 35.4 | 34 | **70.8** | 1 | 2.1 |
| MTPE | 19 | **39.6** | 13 | 27.1 | 7 | 14.6 | 9 | 18.8 | 29 | **60.4** | 0 | 0.0 |
| MT | 38 | **79.2** | 6 | 12.5 | 1 | 2.1 | 2 | 4.2 | 9 | **20.8** | 1 | 2.1 |

Table 6: Reproduction and CS in the three translation modalities

The results show that MT has the fewest CSs (HT has 3.78 times more CSs than MT; and MTPE has 3.22 times more CSs than MT), while HT has the highest number of CSs (1.17 times more than MTPE).

If the type of shifts is observed, HT is clearly above MTPE when it comes to Concretisation, where translators made the TT more detailed, explicit or concrete than the ST. This may be a manifestation of the widely-discussed explicitation strategy in translation (Vinay and Darbelnet 1958); it is also in line with recent research in MT post-editing that finds that post-edited units are simpler and more normalised and have a higher degree of interference from the ST than HT (Toral 2019; Vanmassenhove, Shterionov, and Way 2019). As the translators said in their post-translation questionnaire when asked about post-editing: "my creativity was limited", "I found it more difficult to think outside the box", "The translator feels the text as not fully their own".

*Creativity Score.* Table 7 shows the creativity scores for each translation modality according to the formula presented earlier.

| TM | HT | MTPE | MT |
|---|---|---|---|
| # CS | 34 | 29 | 9 |
| Units | 48 | 48 | 48 |
| # Errors | 49 | 45 | 173 |
| TT words | 2371 | 2369 | 2364 |
| Creativity Score | 68.77% | 58.52% | 11.43% |

Table 7 Creativity scores according to translation modality

HT shows a higher creativity score than MTPE, a finding that is at odds with previous research that found no difference in creativity between translation modalities (Vieira et al. 2020); the divergent findings here may be related to the different methodologies used. This score seems to confirm the view that MT is less creative than a translation done or post-edited by translators, since it has not only more errors but also fewer creative shifts. HT also scores highest for creativity, coinciding with the view from the translators that they are more creative when working on their own than when post-editing.

It is important to reiterate at this point that creativity in this study encompasses acceptability and novelty. There is agreement in creativity studies, in general, and in creativity in translation, in particular, that creativity is characterized by two concepts: novelty (originality) and value (usefulness, appropriateness and effectiveness) (Runco and Jaeger 2012; Dunne 2017). A high number of errors does compromise the translation usefulness and, hence, errors should be considered in this creativity score. Otherwise, anybody, not only translators, could generate several CSs (departing from the source text) while at the same time not reflecting, for example, the meaning of the source text.

*Results on User experience*

*Narrative Engagement*

Figure 2 and Table 8 show overall engagement, which is given by the mean value of all 12 items in the scale (N= 88).

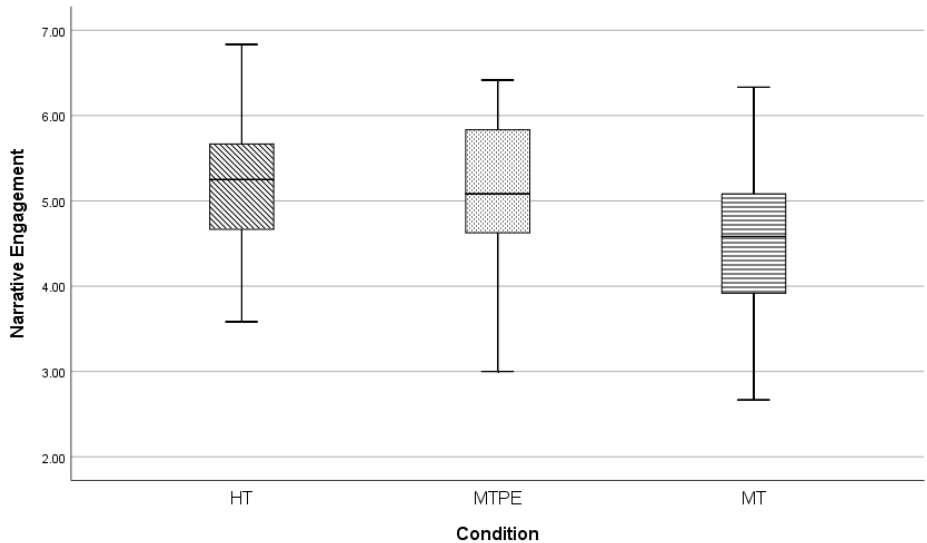

Figure 2. Narrative Engagement according to translation modality

| TM | Mean | Median | STD | Min | Max | Range |
|---|---|---|---|---|---|---|
| **HT** | 5.21 | 5.25 | 0.29 | 3.58 | 6.83 | 3.25 |
| **MTPE** | 5.07 | 5.08 | 0.95 | 3.00 | 6.42 | 3.42 |
| **MT** | 4.61 | 4.58 | 0.98 | 2.67 | 6.33 | 3.67 |

Table 8 Descriptive values for Narrative Engagement according to translation modality

The Cronbach's alpha reliability coefficient is 0.76, which is considered a good reliability score for the items in the scale.[14] A Kruskal-Wallis H test[15] was performed to explore the Narrative Engagement scores according to the translation modality. There is a statistically significant difference between modalities (H(2) = 6.17, p = 0.05) with a mean rank score[16] of 50.26 for HT, 47.43 for MTPE and 33.94 for MT. Post-hoc comparisons using the Mann-Whitney[17] test show statistically significant differences between HT and MT (U= -2.39; p = .02) and MTPE and MT (U= -1.95; p = .05), but not between MTPE and HT.

---

[14] The Cronbach's alpha is a test of internal consistency of items in a scale, i.e. if the items measure what they should measure.

[15] This test is used to determine if there are statistically significant differences between two or more groups within the independent variable (translation modality) when the scale uses rank-based nonparametric values (in our case a 7-point Likert scale).

[16] The mean rank is the average of ranks for all observations with each sample. In this case, the higher value indicates a higher engagement.

[17] The Mann-Whitney test is used to determine the differences between the three groups (HT, MTPE and MT) within the independent variable (translation modality) when the dependant variable is ordinal (rank).

An independent analysis on the four categories of Narrative Engagement reveals that only narrative understanding shows statistically significant differences between translation modalities ($H(2) = 14.31$; $p < 0.01$) with a mean rank of 51.25 for HT, 50.70 for MTPE, and 28.26 for MT. Post-hoc comparisons using the Mann-Whitney test show statistically significant differences between HT and MT ($U= -3.04$; $p < 0.01$) and MTPE and MT ($U= -3.56$; $p < 0.01$), but not between MTPE and HT. Attentional Focus, Narrative Presence, and Emotional Engagement show no statistically significant differences between modalities.

Narrative understanding refers to the ease of comprehending a narrative, as "… audience members should be unaware when comprehension progresses smoothly, and become aware only when comprehension falters" (Busselle and Bilandzic 2009: 341). It appears that MT makes understanding the story more difficult. However, the other categories, namely Attentional Focus (the state of being engaged and not distracted), Emotional Engagement (feeling for and with the characters) and Narrative Presence (the feeling that one has entered the world of the story) are not significantly affected by MT. We conjecture that this could partially be due to the genre and type of story that the readers were exposed to, and that this might be different with a story that is not as dramatic and emotional as the one presented here.

To complement this result, we checked the number of correct answers from the survey (see Research Methodology) and the translation modality, but in this case, there were no statistically significant differences. Therefore, translation modality did not affect participants when responding to the comprehension questions, although they perceived that making sense of the story, and understanding the characters and the thread of the story differed according to the translation modality.

*Enjoyment*

Figure 3 and Table 9 show the overall enjoyment, which is given by the mean value of 3 items in the scale (N= 88).

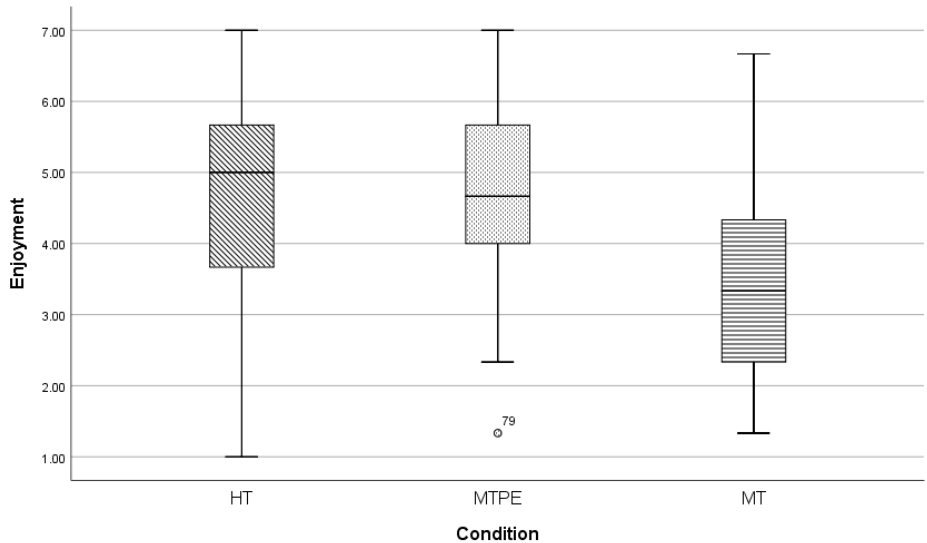

Figure 3 Enjoyment according to translation modality

| TM | Mean | Median | STD | Min | Max | Range |
|---|---|---|---|---|---|---|
| **HT** | 4.60 | 5.00 | 1.52 | 1.00 | 7.00 | 6.00 |
| **MTPE** | 4.64 | 4.67 | 1.33 | 1.33 | 7.00 | 6.00 |
| **MT** | 3.37 | 3.33 | 1.45 | 1.33 | 6.67 | 5.33 |

Table 9 Descriptive values for Enjoyment according to translation modality

The outlier in Figure 3 (participant 79 in MTPE) did not give any comments as to why she disliked the story (M = 1.33). Her mean translation reception score was above average (4). It appears, therefore, that she did not enjoy the story rather than the translation.

The Cronbach's alpha reliability coefficient is 0.84, which is considered a good reliability score for the items in the scale. A Kruskal-Wallis H test was performed to explore the Enjoyment scores according to the translation modality. There is a statistically significant difference between translation modalities (H(2) = 12.51, $p < 0.01$) with a mean rank score of 50.59 for MTPE, 50.52 for HT and 29.24 for MT. Post-hoc comparisons using the Mann-Whitney test show statistically significant differences between HT and MT (U= -2.99; $p < 0.01$) and between MTPE and MT (U= -3.22; $p < 0.01$), but not between MTPE and HT.

Unsurprisingly, MT scores are the lowest once again, but what is somewhat surprising is that HT and MTPE show very similar scores in all questions. Again, we can see that MTPE is a solution that readers enjoy marginally more than (or at least as much as) HT, even if they engaged (according to the Narrative Engagement scale) more with HT, although the difference was not significant.

*Translation Reception*

There were 4 items in the Translation Reception scale. The Cronbach's alpha coefficient for this scale was 0.44. The 4th item "How would you like to read a text by the same author but a different translator?" was not consistent with the other elements. The reason being that it is strongly related to "How would you like to read a text by the same author and translator?", i.e., if readers select that they are likely to read a text by the same author and translator, it is highly probable that they select that they are unlikely to read a text by the same author and a different translator. This internal "contradiction" is problematic for the scale. Therefore, we calculated the reliability coefficient without this 4th question and obtained a 0.71 coefficient. Figure 4 and Table 10 show translation reception, which is given by the median value of 3 items related to translation reception (N= 88).

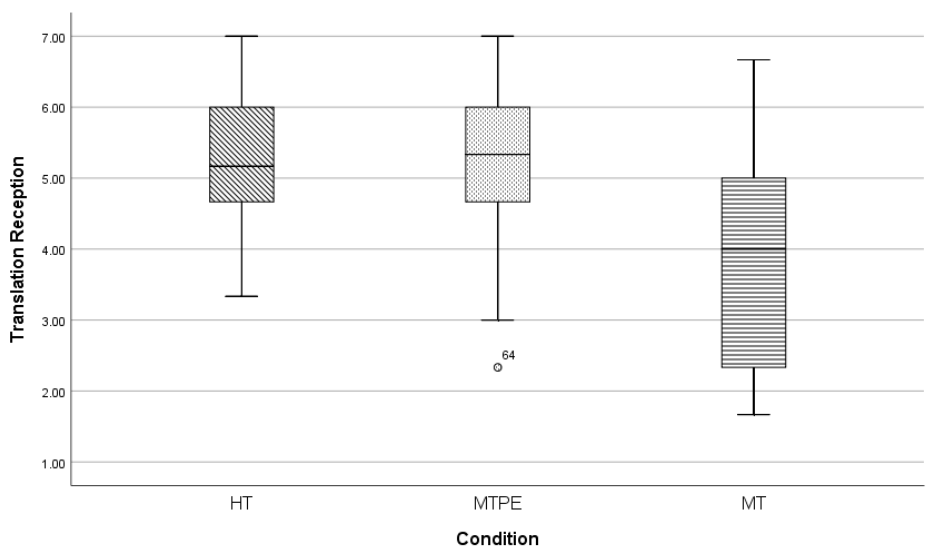

Figure 4 Translation reception according to translation modality

| TM | Mean | Median | STD | Min | Max | Range |
|---|---|---|---|---|---|---|
| **HT** | 5.27 | 5.17 | 1.06 | 3.33 | 7.00 | 3.67 |
| **MTPE** | 5.17 | 5.33 | 1.12 | 2.33 | 7.00 | 4.67 |
| **MT** | 3.85 | 4.00 | 1.51 | 1.67 | 6.67 | 5.00 |

Table 10 Descriptive values for Translation Reception according to translation modality

The outlier in Figure 4 (participant 64 in MTPE) did not give any comments to explain why she disliked the translation (M = 2.33), but, surprisingly, she stated that she did not realize that it was a translation. Her enjoyment was also low (M = 2.33) however. It appears that she did not like the story and found the text difficult to understand.

A Kruskal-Wallis H test was performed (with 3 items) to explore the Translation Reception scores according to the translation modality. There is a statistically significant difference between translation modalities (H(2) = 13.86, p < 0.01) with a mean rank score of 51.68 for HT, 50.19 for MTPE and 28.50 for MT. Post-hoc comparisons using the Mann-Whitney test show statistically significant differences between HT and MT (U= -3.31; p < 0.01) and between MTPE and MT (U= -3.23; p < 0.01), but not between MTPE and HT.

Participants were also asked if they had noticed that they were reading a translation or if there were paragraphs that were difficult to read or if there were paragraphs that they particularly liked. The participants could respond "Yes", "No" or "I don't know". MT shows the highest percentage of readers who were aware of the translation (HT = 57.1%, MTPE = 57.1% and MT =88%). MT shows also the highest percentage for paragraph difficulty (HT = 39.3%, MTPE = 34.3% and MT = 64%) and the lowest for paragraphs liked (HT = 60.7%, MTPE = 42.9% and MT = 36%). When the readers commented on the translations and paragraphs, it was clear that they found the MT output confusing and "too literal", although MT was never mentioned explicitly in the comments (the participants did not know the condition/modality they were reading, but we thought that some of the translations in the MT condition would alert participants of the source of the translation).

*Correlations*

To test if the variable Translation Reception is related to Narrative Engagement or Enjoyment, a correlation analysis was run. A Spearman correlation shows a statistically significant correlation ($r_s(88) = 0.50$; p < 0.01), considered to be moderate, between translation reception and Narrative Engagement, and a statistically significant correlation ($r_s(88) = 0.68$; p < 0.01), considered moderate to strong, between Translation Reception and Enjoyment. In other words, participants who liked the translation better were also more engaged and enjoyed the text more.

*Conclusions*

This study addressed two research questions. The first one was "Can we quantify the creativity in texts translated by humans as opposed to those produced with the aid of machines?"

The method applied to this study, by quantifying errors and creative shifts, gives us an interesting perspective on the differences between translation modalities, and reveals that HT scores higher for creativity than MTPE, all other things being equal. Clearly, the two modalities that show the highest creativity scores are those where professional translators intervene. Translators' opinions on how post-editing constrains their creativity are also proffered in our experiment. These will be evaluated in further studies.

Our creativity analysis can be adjusted and improved. The analysis should be done by more than one expert reviewer and it could be more exhaustive, not only using primary shifts, but possibly also secondary shifts (metaphors, lexical richness, rhythm, for example). Furthermore, it is difficult to come up with a creativity score that includes acceptability and novelty in a single formula that makes sense mathematically and that considers the entire text. We have attempted to create one in this study but are aware of its limitations, i.e. we have assigned points to CSs and errors, but we are aware that in this formula, errors do not have as much weight as CSs because the full target text as denominator is larger than the number of units of creativity potential.

The second question was "Do users exposed to different translation modalities have different reading experiences?". The clear answer is yes. The reading experience, if we consider that narrative engagement, enjoyment and translation reception, are different depending on the modality. HT scores higher in narrative engagement and translation reception and is slightly lower than MTPE in enjoyment. However, there are no statistically significant differences between HT and MTPE for any of these variables. MT, unsurprisingly, has the lowest engagement, enjoyment and translation reception scores. It is noteworthy, though, that those categories related to attentional focus, emotional engagement and narrative presence do not show statistically significant differences across the modalities. This could be related to the nature of the story readers were presented with, as discussed earlier. This seems to hint at the possibility that contemporary MT might be able fulfil a communicative function for some types of literary texts even if the reading experience is not as optimal as with the other modalities where translators intervene. To come to sounder conclusions, it will be necessary to test these variables not only in different language combinations, but also with different genres and literary styles.

The overall question driving our research is *How does creativity in different translation modalities impact the reading experience?* We consider it plausible that creativity in different translation modalities could be the factor that heightens the reading experience. We have seen here that, although HT and MTPE show similar reading experiences, HT ranked higher in all narrative engagement categories, and this is an interesting trend to explore with different texts and languages. The creative shift analysis reveals that the translators in HT provide a more novel translation, less constrained by the MT output, and as a result this might increase narrative engagement. However, it is also clear from the results that readers enjoyed MTPE marginally more or as much as HT, even if they appreciate that HT was a better translation, but this again could differ depending on the genre or style of the story analysed. Finally, the correlation analysis shows a clear link between translation reception and the readers' engagement and enjoyment of a story. What seems to be clear is that professional translators add the creativity factor, by providing solutions that are both novel and acceptable, that MT is lacking at present.

*Funding information*

This project has received funding from the European Union's Horizon 2020 research and innovation programme under the Marie Skłodowska-Curie grant agreement No 890697 and from from CLCG's 2020 budget for research participants and has been partially funded by the Expanding Excellence in England Programme funded by Research England.

*Acknowledgements*

We would like to thank the translators: Marta Pera Cucurell, Josep Marco Borillo and Núria Molines Galarza for the invaluable contributions and expertise, as well as the participants in the survey.